\pdfoutput=1
\documentclass{IEEEtran} 
\usepackage{mathtools} 
\usepackage{cite} 
\usepackage{soul}
\usepackage{graphicx} 
\usepackage[outdir=./]{epstopdf}
\usepackage{array} 
\usepackage{amsthm} 
\usepackage{amsfonts}
\usepackage{multirow} 
\usepackage{adjustbox} 
\usepackage{multicol} 
\usepackage{supertabular} 
\usepackage{wrapfig} 
\usepackage{lipsum}
\usepackage{footnote} 
\usepackage[absolute]{textpos} 
\usepackage{threeparttable} 
\usepackage{empheq}
\usepackage{colortbl}
\usepackage{flexisym} 
\usepackage{xcolor} 
\usepackage{tablefootnote} 
\usepackage{blindtext} 
\usepackage{multirow} 
\usepackage{spreadtab} 
\usepackage{amsmath} 
\usepackage{mathtools} 
\usepackage[normalem]{ulem} 
\usepackage{array} 
\usepackage{soul}
\usepackage{pifont}
\usepackage[linesnumbered,lined,boxed,commentsnumbered,ruled]{algorithm2e}
\usepackage{caption}
\usepackage{subcaption}
\usepackage{booktabs}
\usepackage{hyperref}
\definecolor{light-gray}{gray}{0.9}
\usepackage{soul}

\newcolumntype{P}[1]{>{\centering\arraybackslash}p{#1}} 
\newcolumntype{M}[1]{>{\centering\arraybackslash}m{#1}} 
\usepackage[version-1-compatibility]{siunitx} 
\theoremstyle{plain}

\theoremstyle{definition}

\theoremstyle{remark}

\title{Fixflow: A Framework to Evaluate Fixed-point Arithmetic in Light-Weight CNN Inference}
\author{{Farhad~Taheri,
		Siavash~Bayat-Sarmadi,~\IEEEmembership{Member,~IEEE}, Hatame~Mosanaei-Boorani, Reza~Taheri 
	}
	\thanks{All authors are with the Department of Computer Engineering, Sharif University of Technology, Tehran, Iran. 
		Emails: f.taheri89@sharif.edu, mosanaei@ce.sharif.edu, rezataheri@sharif.edu sbayat@sharif.edu. 
		This publication was supported by grant No. RD-51-9911-0030 from the R\&D Center of Mobile Telecommunication Company of Iran (MCI) for advancing information and communications technologies.}}

\begin{document} 
	\maketitle
	\begin{abstract}
		Convolutional neural networks (CNN) are widely used in resource-constrained devices in IoT applications. In order to reduce the computational complexity and memory footprint, the resource-constrained devices use fixed-point representation. This representation consumes less area and energy in hardware with similar classification accuracy compared to the floating-point ones. However, to employ the low-precision fixed-point representation, various considerations to gain high accuracy are required. Although many quantization and re-training techniques are proposed to improve the inference accuracy, these approaches are time-consuming and require access to the entire dataset. This paper investigates the effect of different fixed-point hardware units on CNN inference accuracy.
		To this end, we provide a framework called Fixflow to evaluate the effect of fixed-point computations performed at hardware level on CNN classification accuracy. We can employ different fixed-point considerations at the hardware accelerators.This includes rounding methods and adjusting the precision of the fixed-point operation's result.
		Fixflow can determine the impact of employing different arithmetic units (such as truncated multipliers) on CNN classification accuracy. Moreover, we  evaluate the energy and area consumption of these units in hardware accelerators. We perform experiments on two common MNIST and CIFAR-10 datasets. Our results show that employing different methods at the hardware level specially with low-precision, can significantly change the classification accuracy.		
		
	\end{abstract}



\section{Introduction}
\label{sec:introduction}
\setlength{\extrarowheight}{1mm}
Deep learning is an outstanding solution to solve different problems in a wide range of applications.
Convolutional neural networks (CNNs) are among the most emerging deep learning algorithms that provide remarkable accuracy for classification tasks. Recently, the accuracy of CNNs is comparable or even better than humans; therefore, CNN is employed in many applications such as image classification~\cite{krizhevsky2012imagenet}, speech processing~\cite{sainath2013deep}, and robotics~\cite{levine2016end}. Despite the high classification accuracy, CNNs have high computation complexity~\cite{sze2017efficient, taheri2022risc}, e.g., 30 billion operations (multiplication and addition) should be performed to classify a single image~\cite{lian2019high}. To overcome this challenge, previous studies employ GPU, field-programmable gate array~(FPGA), or application-specific integrated circuit~(ASIC) platforms to accelerate CNNs~\cite{sze2017efficient, chen2016eyeriss, parashar2017scnn, chetlur2014cudnn, lavin2016fast}. Although GPUs have sufficient computational power, they have high energy consumption. Therefore, GPUs are not suitable for IoT or resource-constrained applications and usually are used for training~\cite{parashar2017scnn, gao2017tetris}. In previous works, FPGA/ASIC platforms are used to accelerate CNN inference on edge devices due to their energy efficiency and configurability~\cite{garofalo2020xpulpnn, gao2017tetris}. 

In order to improve the performance and energy consumption in FPGA/ASIC accelerators for a CNN inference, previous works quantize the floating-point number to a different representation~\cite{song2020drq, liu2020optimize, rastegari2016xnor}. 
In~\cite{garofalo2020xpulpnn, lo2018fixed}, the authors use a low-precision fixed-point arithmetic (e.g., 4-bit, 6-bit) to improve the energy consumption of the accelerators for resource-constrained applications such as IoT. In case of employing a low-precision fixed-point representation in hardware accelerators, the accuracy may decrease. Therefore, different techniques are proposed to improve the accuracy loss in hardware accelerators using low-precision fixed-point. In~\cite{bhalgat2020lsq+}, the fine-tuning and quantization methods are proposed to re-trained the model for fixed-point numbers. The work presented in \cite{gysel2018ristretto, lo2018fixed} proposes dynamic fixed-point numbers, which means using different widths for integer and fraction parts in different layers, to improve the accuracy. Although these approaches can prevent accuracy loss, implementing this method requires a time-consuming pre-processing step, adding complexity to the hardware accelerator and access to the entire dataset, which can cause privacy issues. 

Recent studies propose post-training quantization methods to improve the classification accuracy of low-precision quantization~\cite{krishnamoorthi2018quantizing, kryzhanovskiy2021qpp, banner2019post}. Post-training quantization methods enhance accuracy by computing the quantization parameter, such as the scaling factor at run time. Therefore, these methods do not require retraining and training datasets.

In~\cite{gupta2015deep}, the authors investigate different rounding methods to reduce the precision for the training phase. This work shows that with stochastic rounding, the hardware accelerators can use a 16-bit fixed-point number instead of a 32-bit number in the training phase. Similarly, for CNN inference hardware accelerators, using a fixed-point number requires various considerations, including 1) weights and inputs of the CNN should be converted from floating-point to fixed-point representation; 2) after each arithmetic operation on a fixed-point number, the higher precision result should be reduced to the desired precision (i.e., the precision should be adjusted). To this end, we should select the appropriate part of the number after the fixed-point operation and employ a suitable rounding method to remove the excess part of the fixed-point number. 

This paper introduce a framework called Fixflow to investigates the impacts of employing different hardware fixed-point computations on CNN classification accuracy. The aim of this study is to evaluate the hardware configurations on the inference accuracy.
Unlike time-consuming re-training or post-training quantization methods in this work we want to evaluate hardware design configuration on classification accuracy.  
We provide a comprehensive evaluation of different ways to select the desired bits after each arithmetic operation and use different rounding methods. First, we train the model and extract the weights, and then we convert the weights and test set to the fixed-point representation. Finally, the framework measures the inference accuracy for different configurations of the fixed-point representation. 
Our results show that these methods can significantly affect the inference accuracy specially in the low-precision fixed-point representations. For instance, by applying different methods in 4-bit fixed-point representation, the accuracy changes up to 87.74\% on the MNIST dataset without employing any fine-tuning or dynamic representations. Moreover, we propose a hardware implementation for these different rounding methods and provide hardware metrics (area, energy) for hardware accelerators using fixed-point computation. 

The contributions of the paper are summarized as follows:
\begin{itemize}
	\item We developed a framework called Fixflow to evaluate the effect of fixed-point computation at hardware level on CNN accuracy . The source code of Fixflow is available online at \url{https://github.com/3S-Lab/FixFlow}.
	\item We investigate different methods to adjust the precision of the fixed-point operation's result in CNN hardware accelerators.
	\item We evaluate the effect of different rounding methods, which are used to convert floating-point number to fixed-point numbers and remove the excess part of fixed-point numbers, on classification accuracy.
	\item We evaluate different hardware units in Fixflow that can help hardware designers to improve the hardware accelerators in terms of accuracy.
	\item We also provide evaluations of employing these methods on CNN hardware accelerators. 
\end{itemize}

This paper is organized as follows. In Section~\ref{sec:background}, we provide the preliminaries. Section~\ref{sec:methodology} proposes various methods to adjust the precision of fixed-point numbers. In Section~\ref{sec:eval}, we provide the classification accuracy on different datasets regarding each method introduced in Section~\ref{sec:methodology}. Section~\ref{sec:conclusion} discusses our proposed approach to perform fixed-point computation in hardware accelerators. Finally, we conclude this work in Section~\ref{sec:conclusion}.


\section{Background}
\label{sec:background}
Convolutional Neural Networks (CNNs) are a type of Deep Neural Networks (DNNs), which are commonly used for image classification. These networks are constructed with multiple connected layers. There are four kinds of layers used in CNNs: convolutional (Conv), pooling, activation, and fully connected (FC) layers. Additionally, normalization layers are used in some CNNs to improve the training time and accuracy~\cite{krizhevsky2017imagenet}.

\begin{figure}[t]
\centering
\includegraphics[scale=0.4]{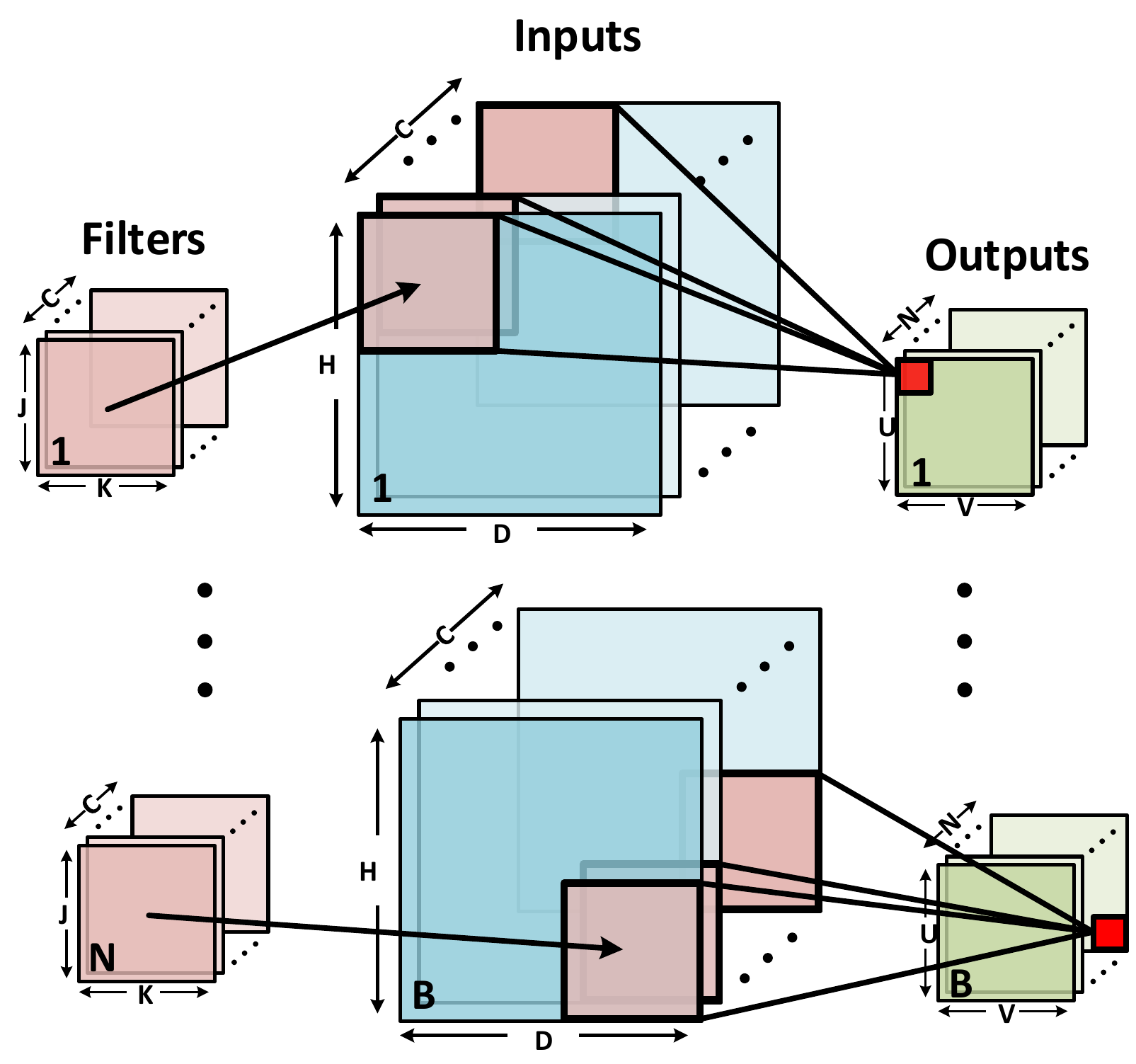}
\caption{Convolutional layer in CNN.}
\label{fig:cnv}
\vspace{-8pt}
\end{figure}

Conv layers extract features of the inputs by convolving the 3-D filter weights with the inputs. Fig.~\ref{fig:cnv} shows the Conv layers in CNN. Pooling layers are used to reduce the size of the inputs. Pooling layers, replace the numbers in a 2$\times$2 or 3$\times$3 window with the maximum or average value of the window. Activation layers apply a non-linear function on the output of Conv and FC layers. ReLU is a typical activation function used in CNNs due to their simplicity and for improving training time. FC layers are typically used as the last layers of CNNs to classify the input. This layer is similar to the Conv layer, but the filter window size equals the input (no weight sharing in Conv layers). The main computational cost of CNNs is related to the multiply accumulation (MAC) operations, which are used in Conv and FC layers.


\section{Related Work}
\label{sec:related work}

In spite of having an outstanding performance in classification, CNNs suffer from high computational complexity. Therefore, previous works propose various hardware accelerators to accelerate the computation of CNNs~\cite{chen2014diannao, chen2016eyeriss}. Recent studies improve the efficiency of hardware accelerators by employing different method such as quantization or pruning the unnecessary computations in a CNN~\cite{albericio2016cnvlutin, parashar2017scnn, li2019squeezeflow, kundu2020pre}. Floating-point is a common representation that is used in training and inference phases of CNNs on CPU/GPU. However, related works show that the floating-point representation is not an efficient one for FPGA/ASIC accelerators; thus, many quantization techniques are proposed to reduce energy, area and memory overhead in FPGA/ASIC accelerators~\cite{vogel2018efficient, song2020drq, qin2020binary}. The work presented in \cite{horowitz20141} shows that compared to a floating-point multiplication unit, an 8-bit fixed-point multiplier improves energy consumption by 18.5$\times$ and area by 116$\times$, with a negligible accuracy loss. In \cite{liu2020optimize, vogel2018efficient}, authors employ the logarithmic representation in hardware accelerators. In this representation, the multiplication can be replaced with a shift operation. 
Moreover, a more aggressive quantization approach called binary is provided in~\cite{rastegari2016xnor, umuroglu2017finn, qin2020binary, zeng2020addressing}. In the binary representation, the multiplication operation can be replaced with an XNOR. However, these approaches achieve efficiency by reducing the classification accuracy. 

Fixed-point representation is commonly used in many CNN accelerators due to their negligible accuracy loss and low energy consumption compared to the floating-point representation. Low-precision fixed-point numbers can be used to improve the hardware efficiency at the cost of accuracy loss.
To overcome this challenge, authors in \cite{gysel2018ristretto} propose a framework for dynamic fixed-point representation. This framework can change the bit width of integer/fraction for any number in different layers to improve accuracy. The dynamic precision fixed-point representation is introduced in order to use the high precision fixed-point number in sensitive regions or different layers~\cite{song2020drq, sharma2018bit}. Dynamic precision fixed-point representation can improve the accuracy but implementing these methods on hardware is challenging.
Hashemi et al.~\cite{hashemi2017understanding} proposes a comprehensive evaluation on the impact of different fixed-point precisions on the classification accuracy and energy consumption of CNN accelerators. However, they do not consider various methods of performing fixed-point computation, i.e., rounding methods. Gupta et al.~\cite{gupta2015deep} proposed and evaluated the impact of various rounding methods on training a CNN with fixed-point representation. This work applies the stochastic rounding (SR) for the training phase. They show that the model can be trained with lower precision (16-bit instead of 32-bit) by using SR. However, the impact of various rounding methods on the CNN inference accuracy is not considered in this work. Moreover, this work does not propose any hardware implementations for rounding methods.  
In~\cite{krishnamoorthi2018quantizing, kryzhanovskiy2021qpp, banner2019post}, the authors propose a post-training quantization method to improve the inference accuracy on low-precision quantization without time-consuming re-training methods. However, these methods explore a scaling factor for weights and input features for each layer and do not consider hardware design parameters on classification accuracy.


\section{Methodology}
\label{sec:methodology}
\begin{figure*}[htb]
	\centering
	\includegraphics[scale=0.88]{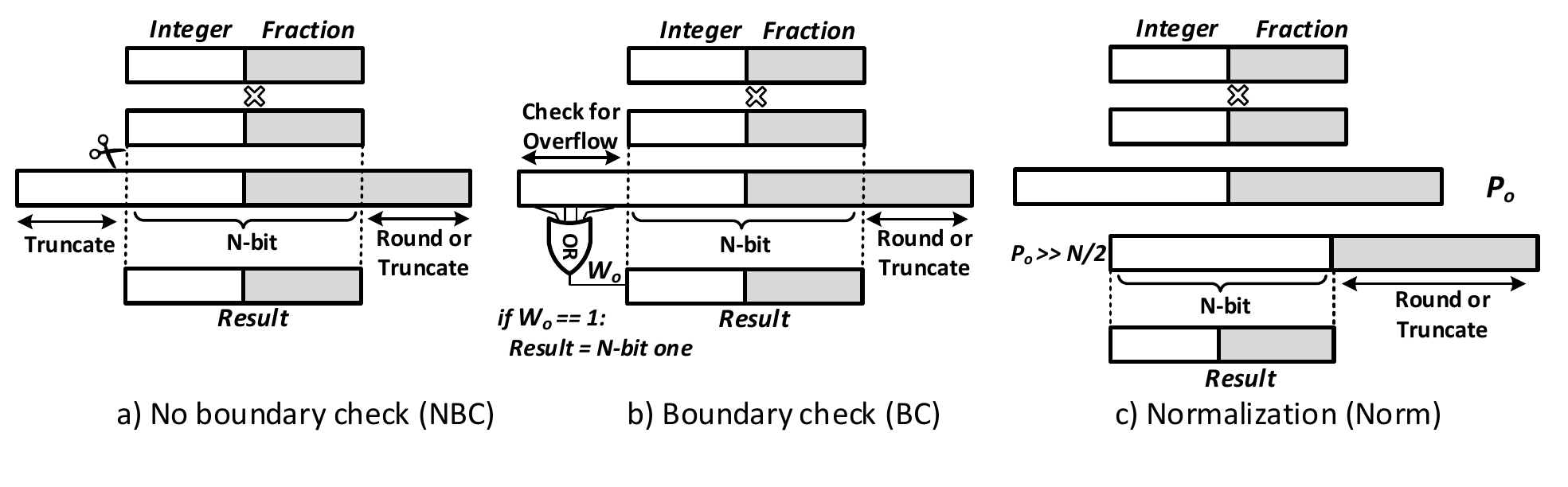}
	\caption{Selecting methods for fixed-point multiplication.}
	\label{fig:select}
\end{figure*}

This section investigates various approaches to perform fixed-point arithmetic in hardware accelerators that are supported in Fixflow. First, we discuss approaches to reduce the higher precision result after each fixed-point operation, such as multiplication (i.e., adjusting the precision of fixed-point operation's result). Next, we introduce rounding methods that can be used to remove the excess part of the fixed-point number. Finally, the above considerations can be performed at different positions in the hardware accelerators (i.e., after multiplication or after multiplication-accumulation (MAC) operations); hence, we discuss the impact of these positions in the hardware accelerators.

\subsection{Adjusting Fixed-point Precision}
\label{subsec:operation}
In the fixed-point representation after each operation, the bit width of the result is extended; therefore, we can employ several approaches to adjust the precision of fixed-point after each operation. To the best of our knowledge, the impact of using these approaches has not been evaluated in CNN inference hardware accelerators. For the sake of simplicity, in this section, we focus on the multiplication operation. Note that these approaches can also be used in converting floating-point numbers to the fixed-point representation. 

As shown in Fig.~\ref{fig:select}, multiplying two N-bit numbers produces a 2N-bit result in the fixed-point representation. Thus, after each multiplication, we should reduce the precision from 2N-bits to N-bits. Fig.~\ref{fig:select} depicts different methods used to adjust the precision of the multiplication's result. For LSBs of the result, we can use different rounding or truncating methods, which are discussed in Section~\ref{subsec:rounding}. Here, we discuss methods to select N-bits from a 2N-bit number. In the following, we explain three different selecting methods.

\textbf{No boundary check (NBC):} Fig.~\ref{fig:select}a shows the NBC method. As shown in the figure, after performing a multiplication, the middle part of the result is selected, and MSBs of the result are clipped without any consideration. For LSBs of the result, we can use any rounding or truncating method that is introduced in Section~\ref{subsec:rounding}. The NBC method can lead to a huge accuracy drop in low-precision fixed-point numbers (e.g., 4-bit, 6-bit) because MSBs of the result are clipped without any consideration.

\textbf{Boundary check (BC):} In the boundary check method, if we encounter an overflow in the result, the multiplication's result gets clipped to the maximum or the minimum number represented with N-bits fixed-point. Otherwise, the BC method behaves similarly to the NBC method. The BC method can be implemented with OR gates in the hardware to check the occurrence of overflow. Fig.~\ref{fig:select}b shows the BC method. This approach reduces the accuracy loss with a negligible hardware cost. For instance, in the NBC method, if we have a 1-bit overflow and the other bits of the result are equal to zero, the final result equals zero. While in the BC method, the final result is equal to the maximum value shown by an N-bit fixed-point number, which is much closer to the correct result.

\textbf{Normalization (Norm):} This method normalizes the multiplication's result to a lower fixed-point representation (N-bit) similar to post-training quantization. To this end, the result of a multiplication operation has to be divided by a power-of-two number. This operation is implemented by shifting or selecting the N most significant bits of the result in the hardware. Next, for LSBs of the result, we can apply the rounding or truncating method. Note that, to achieve a high classification accuracy with this approach, we should follow some considerations; otherwise, the classification accuracy drops significantly even for high precision numbers. 
Since this approach selects MSBs of the result, inputs and weights should be scaled to prevent MSBs from becoming zero. For instance, to represent the weights in the range of [-1, 1] with an 8-bit fixed-point number (with a 4-bit integer), MSBs of the weights always becomes zero. Therefore, after each multiplication, MSBs of the result also becomes zero, and we face a significant accuracy loss. In order to overcome this issue, we should correctly scale the inputs and weights to a desired fixed-point precision.

\subsection{Rounding Methods}
\label{subsec:rounding}
In previous work, the use of rounding methods has only been investigated in the training phase of CNNs~\cite{gupta2015deep}. Furthermore, truncation is a common method that is used to reduce the fixed-point precision in CNN accelerators~\cite{chen2016eyeriss, lo2018fixed}. To the best of our knowledge, the impact of rounding methods on classification accuracy and implementation complexity has not been investigated yet. Rounding methods can be used to reduce the accuracy loss, especially in low-precision fixed-point numbers, e.g., 4-bit or 6-bit numbers, with a negligible hardware complexity.
In CNN hardware accelerators, rounding methods can be potentially used in two scenarios: 1) converting the floating-point number (weights, bias, and inputs) into a fixed-point representation; 2) converting a higher precision fixed-point number to a lower precision number after each fixed-point operation. In this work, we investigate the impact of using three prevalent rounding methods on classification accuracy: stochastic rounding (SR), round-to-nearest (RN), and ROM rounding~\cite{kuck1975rom}. Moreover, we evaluate and compare the truncation method with rounding methods.

Here, $x$ is a number and $\lfloor x \rfloor$ is the target fixed-point representation of $x$. 
$\lfloor x \rfloor$ shows the fixed-point number with \mbox{$(IN, FR)$}, where $IN$ and $FR$ show the integer and fraction width of fixed-point number, respectively. The $\epsilon$ indicates the smallest number ($2^{-FR}$) represented by $\lfloor x \rfloor$. In the following, we explain the rounding and truncating methods:

\textbf{Stochastic rounding:} To perform SR on $x$, first, the difference of $x$ and $\lfloor x \rfloor$ should be calculated. This difference indicates the excess bits of $x$ that should be removed. By dividing the difference of $x$ and $\lfloor x \rfloor$ by $\epsilon$, a number is calculated in the range of [0, 1]. Then, by comparing this value with a random number (R) $\in [0, 1)$, the result of SR is determined. Due to the random nature of R, the result of this method is determined. SR is computed as follows:
\begin{equation}
SR(x) = \left\{\begin{array}{lll}
\lfloor x \rfloor & \mbox{if}
& R\geq \dfrac{x -\lfloor x \rfloor}{\epsilon} \\ \lfloor x \rfloor + \epsilon & \mbox{if}
& R <\dfrac{x - \lfloor x \rfloor}{\epsilon}
\end{array}\right.
\end{equation}

\textbf{Round-to-nearest:} In this rounding method, the result is the closest fixed point number to $x$. If the excess bits of $x$ is larger than $\epsilon / 2$, the $\lfloor x \rfloor$ is added by $\epsilon$; otherwise, the result is equal to $\lfloor x \rfloor$. The main disadvantage of RN is that calculating RN requires the add operation.
RN is computed as follows:
\begin{equation}
RN(x) = \left\{\begin{array}{lll}
\lfloor x \rfloor & \mbox{if}
&\lfloor x \rfloor \leq  x < \lfloor x \rfloor + \dfrac{\epsilon}{2} \\ \lfloor x \rfloor + \epsilon & \mbox{if}
& \lfloor x \rfloor + \dfrac{\epsilon}{2} \leq x < \lfloor x \rfloor + \epsilon
\end{array}\right.
\end{equation}

\textbf{ROM rounding:} ROM rounding has been proposed to remove the add operation. In this method, if LSB of $\lfloor x \rfloor$ is equal to 0, this method follows the same procedures as RN. Otherwise, if LSBs of $\lfloor x \rfloor$ is equal to 1, the $\lfloor x \rfloor$ remains unchanged. The ROM rounding has less hardware complexity and can be implemented with a look-up table.

\textbf{Truncating:} This method removes the excess part of the $x$.  Truncating is a naive method commonly used in related works due to the lack of hardware complexity. However, it can lead to an accuracy drop for low-precision fixed-point numbers.
We provide a comprehensive evaluation of using these rounding methods on the classification accuracy in Section~\ref{sec:eval}. Note that, in our experiments, we perform these rounding methods before applying the mentioned selecting methods (\ref{subsec:operation}) to reduce the accuracy loss.

\subsection{Adjusting Position}
\label{subsec:rp}
In hardware accelerators, we can adjust the precision after performing the following two operations: (1) multiplication operation (RMULT); (2) MAC operation (RMAC).
For instance, for an 8-bit fixed-point representation, in the first approach (RMULT), after performing multiplication, we use methods that are introduced in Section~\ref{subsec:operation} to adjust the precision. Next, results can be accumulated with an 8-bit adder. While, in the second approach (RMAC), the multiplication results are accumulated with a 16-bit adder; then, the results are reduced to 8-bit precision. Although, RMAC is more resource-consuming than RMULT, it leads significant improvements in classification accuracy compared to RMULT for low-precision numbers. 

\begin{figure}[t]
\centering
\includegraphics[scale=0.8]{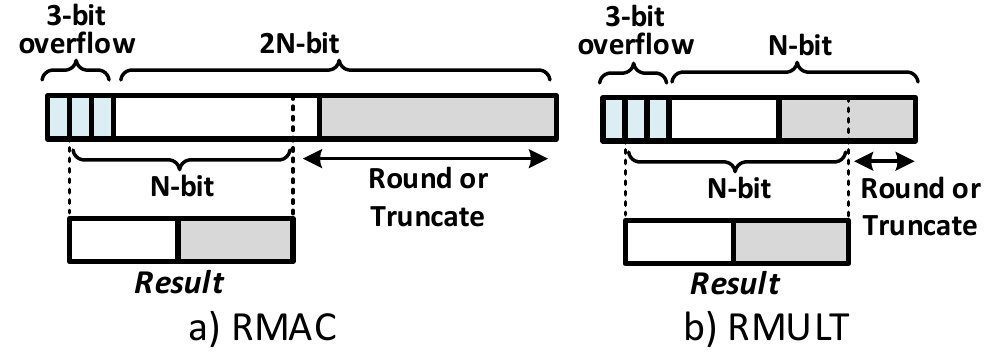}
\caption{Selecting MSBs with the Norm method after last add operation.}
\label{fig:norm}
\end{figure}

Note that these approaches need some consideration while using the Norm method. An overflow usually occurs after the add operation, hence, MSBs of the result should be selected more precisely. In our experiments, we observe that up to 3-bit overflow occurs. Fig~\ref{fig:norm} shows two different approaches for selecting the bits in the Norm method. Our experiments show that if we do not consider overflow or select the entire overflow as a low-precision number, we face a significant accuracy drop. As the 3-bit overflows rarely happen, we select two overflow bits for the final result. With this configuration, the Norm method in RMAC reaches the best classification accuracy compared to other methods. For instance, in 4-bit representation, we achieve 0.6\% accuracy loss compared to the floating-point computation for the MNIST dataset. In Section~\ref{sec:eval}, we provide a comprehensive evaluation of using these methods in hardware accelerators for CNN inference.

\section{Evaluations}
\label{sec:eval}
In this section, first, we introduce an experimental setup, and then we provide accuracy and hardware implementation result for methods discussed in Section~\ref{sec:methodology}.

\subsection{Experimental Setup}
\label{subsec:experimental}
We evaluate 120 different fixed-point configurations proposed in Section~\ref{sec:methodology} in terms of accuracy and design metrics, i.e., area and power. This experiment has been performed on Intel Xeon x5680. In order to evaluate the classification accuracy, due to the lack of supporting fixed-point numbers in common frameworks, we develop a fixed-point framework for CNN inference called Fixflow. Fixflow supports different configurations that are represented in Section~\ref{sec:methodology} including different rounding methods and different fixed-point precision. Moreover, Fixflow can evaluate different arithmetic units, e.g., truncated and approximated multipliers on CNN accuracy. 
To measure the inference accuracy, first, we train the model with Tensorflow~\cite{abadi2016tensorflow} without fine-tuning. Then, the framework converts weights and the test set to a desired fixed-point representation and performs CNN inference by the fixed-point arithmetic. To evaluate Fixflow, we measure the inference accuracy with the framework with floating-point weights and inputs. The inference accuracy obtained from our framework with floating-point numbers is equal to the accuracy obtained from Tensorflow.
 In our experiments, we consider two commonly used image classification datasets, namely MNIST~\cite{lecun1998mnist} and CIFAR-10~\cite{krizhevsky2014cifar}, to evaluate the fixed-point accuracy. 
 For MNIST we use a LeNet-5 architecture~\cite{lecun1998gradient}, and for CIFAR-10 we use an architecture similar to~\cite{krizhevsky2012imagenet}. Since the normalization layer is not implemented in many hardware accelerators, we remove this layer in architecture presented in ~\cite{krizhevsky2012imagenet}. Table~\ref{tab:arch} shows the CNN architectures employed in our experiments. 

For our hardware experiments, we implement DianNao~\cite{chen2014diannao} as baseline architecture and test different configurations on this architecture.
We design and synthesize our work using 45nm NanGate standard cell library~\cite{45nm} with Synopsys Design Compiler~\cite{dc} on the ASIC platform.

\begin{table}[]
\centering
\caption{CNN architectures used in this work}
\label{tab:arch}
\begin{tabular}{c|c}
\cmidrule[1pt](lr){1-2}
\begin{tabular}[c]{@{}c@{}}MNIST use LeNet~\cite{lecun1998gradient}\end{tabular} & \begin{tabular}[c]{@{}c@{}}CIFAR-10 use \cite{krizhevsky2012imagenet}\end{tabular} \\ \cmidrule(lr){1-2}
\begin{tabular}[c]{@{}c@{}}28$\times$28$\times$1\\ Conv 5$\times$5$\times$6\\ Maxpool 2$\times$2\\ Conv 5$\times$5$\times$16\\ Maxpool 2$\times$2\\ FC 84, 10\end{tabular} & \begin{tabular}[c]{@{}c@{}}32$\times$32$\times$3\\ Conv 3$\times$3$\times$32\\ Maxpool 2$\times$2\\ Conv 3$\times$3$\times$64\\ Maxpool 2$\times$2\\ 3$\times$Conv 3$\times$3$\times$128 \\Maxpool 2$\times$2\\ FC 500, 128, 10\end{tabular} \\ \cmidrule[1pt](lr){1-2}
\end{tabular}
\end{table}

\begin{figure*}[t]
\centering
\includegraphics[width=0.9\linewidth]{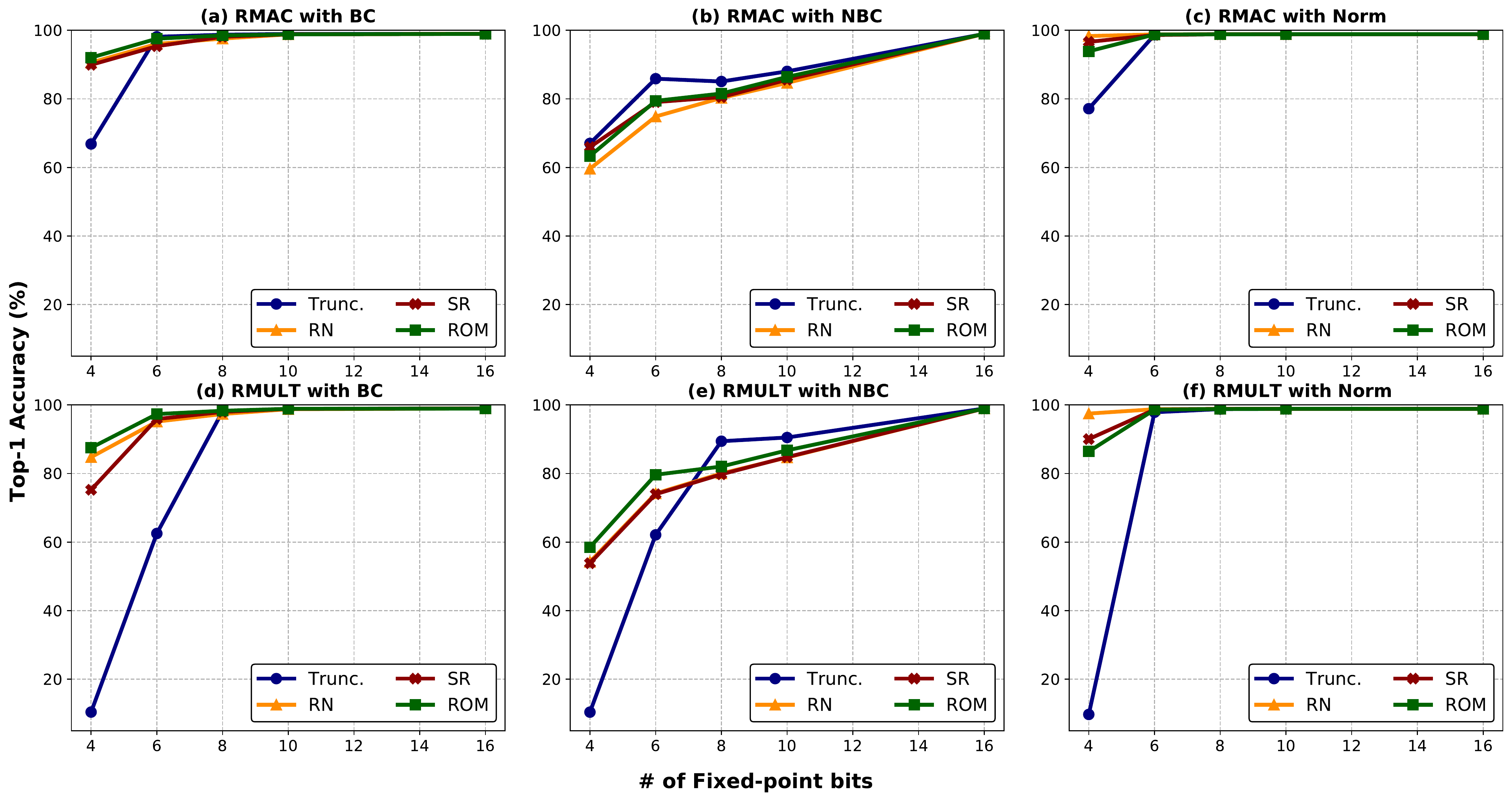}
\caption{Top-1 accuracy for the \textit{MNIST} dataset for different fixed-point lengths. The fixed-point numbers are adjusted to a lower precision after MAC (\textit{top}) and after multiplication (\textit{bottom}). BC (Fig a and d), NBC (Fig. b and e), and Norm (Fig. c and f) and different rounding methods (lines in figures) are used to adjust the precision.}
\label{fig:lenet_res}
\end{figure*}

\begin{figure*}[t]
\centering
\includegraphics[width=0.9\linewidth]{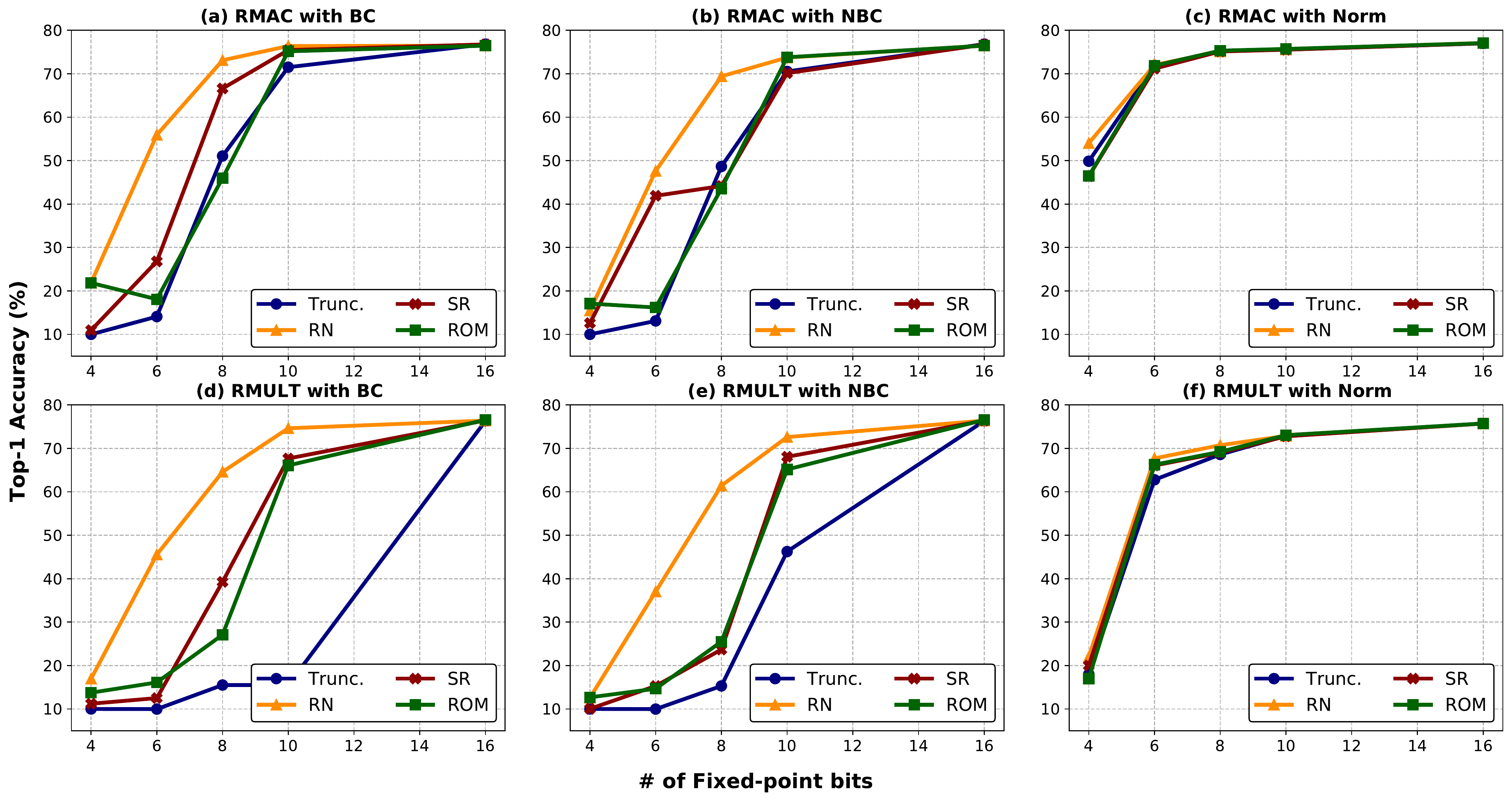}
\caption{Top-1 accuracy for the \textit{CIFAR-10} dataset for different fixed-point lengths. The fixed-point numbers are adjusted to a lower precision after MAC (\textit{top}) and after multiplication (\textit{bottom}). BC (Fig a and d), NBC (Fig. b and e), and Norm (Fig. c and f), and different rounding methods (lines in figures) are used to adjust the precision.}
\label{fig:alexnet_res}
\end{figure*}

\subsection{Results}
\label{subsec:result}
Fig.~\ref{fig:lenet_res} shows the classification accuracy for the MNIST dataset. We employ different methods to adjust the fixed-point precision after MAC and multiplication operations. In our experiments, we achieve 98.9\% accuracy with the floating-point computation. We measure the accuracy with 4, 6, 8, 10, and 16-bit fixed-point numbers. Moreover, as discussed in Section~\ref{sec:methodology}, in the Norm method, we should find the best scaling factor for weights and inputs; otherwise, the classification accuracy drops significantly, but for BC and NBC methods, in all experiments, weights are not scaled, and the inputs are divided by 255. 

We observe from Fig.~\ref{fig:lenet_res} that: (1) In the RMULT approach, the accuracy drops drastically for low-precision, e.g., in 4-bit with the truncation method, the accuracy drops down to 10\% (which literally is random guessing between 0 to 9); (2) RN compared to the truncation method in 4-bit fixed-point representation, increases the accuracy up to 23.74\% for the RMAC approach and 87.74\% for the RMULT approach; (3) The best accuracy for 4-bit fixed-point numbers can be achieved by Norm and RN methods. In this approach, the accuracy drops down to 98.34\% (i.e., 0.6\% accuracy loss compared to the floating-point numbers); (4) For BC and NBC, ROM is a better method compared to RN due to higher overflow occurrence in RN; (5) Unlike training presented in~\cite{gupta2015deep}, the SR method is not an appropriate choice for the inference phase because it has higher hardware overhead and lower accuracy gains than RN and ROM; (6) For more than 10-bit fixed-point numbers, the accuracies obtained from all methods are similar; so the method with the lowest energy consumption is the best choice for high precision fixed-point numbers, e.g., 16-bit representation.

Fig.~\ref{fig:alexnet_res}, shows the classification accuracy on the CIFAR-10 dataset. The baseline accuracy is 76.8\% for the floating-point representation. The configurations are similar to the MNIST dataset. The observations obtained from Fig.~\ref{fig:alexnet_res} indicate that: (1) In a 4-bit fixed-point number, the Norm method in RMAC, unlike~\cite{hashemi2017understanding} reaches to 54.02\% accuracy; however, other approaches fail to gain sufficient accuracy; (2) In an 8-bit fixed-point number, the accuracy drops down to 75.32\% (i.e., 1.48\% accuracy loss compared to the floating-point representation) by using the Norm method; (3) All approaches in 16-bit fixed-point representation have the same accuracy with floating-point representation; (4) Other observations are similar to the MNIST dataset.

\begin{table}[]
	\centering
 \caption{ASIC/FPGA implementation result of single 8-bit multiplication combined with rounding methods.}
\label{tab:rounding_res}
\begin{threeparttable}
\begin{tabular}{c|c|c|c||c|c}
\cmidrule[1pt](lr){1-6}
\multirow{2}{*}[-1.5ex]{\textbf{\begin{tabular}[c]{@{}c@{}} Building\\  Blocks\end{tabular}}} & \multicolumn{3}{c||}{\textbf{ASIC}} & \multicolumn{2}{c}{\textbf{FPGA}} \\ \cmidrule(lr){2-4} \cmidrule(lr){5-6}
 & \textbf{\begin{tabular}[c]{@{}c@{}}Area\\  \textbf{$\mu m^2$}\end{tabular}} & \textbf{\begin{tabular}[c]{@{}c@{}}Power\\  ($\mu$W)\end{tabular}} & \textbf{\begin{tabular}[c]{@{}c@{}}Freq.\\  (MHz)\end{tabular}} & \textbf{LUT} & \multicolumn{1}{c}{\textbf{FF}}  \\ \cmidrule[1pt](lr){1-6}
 \textbf{ Truncation\tnote{$\dag$}} 
 & 437.03 & 9.02 & 869.56 & 70 & 8    \\ \cmidrule(lr){1-6}
 \textbf{ SR} 
 & 601.42 & 13.05 & 819.67 & 82 & 16    \\ \cmidrule(lr){1-6}
 \textbf{ RN} 
 & 576.15 & 13.08 & 833.33 & 78 & 8    \\ \cmidrule(lr){1-6}
  \textbf{ ROM}
 & 440.49 & 9.53 & 869.56 & 71 & 8    \\\cmidrule[1pt](lr){1-6}
\end{tabular}
\begin{tablenotes}
    \item[$\dag$] 8 MSBs of the multiplication result are selected. 
  \end{tablenotes}
 \end{threeparttable}
\end{table}

Table~\ref{tab:rounding_res} shows the hardware implementation result of a single 8-bit multiplication combined with different rounding methods on FPGA and ASIC platforms. We implement these circuits on FPGA with Vivado design suit 2019~\cite{vivado} and ASIC using 45nm NanGate standard cell library~\cite{45nm} with design compiler~\cite{dc}. On the FPGA platform, we synthesize our design without a DSP. As shown in Table~\ref{tab:rounding_res}, RN increases the area by 1.31$\times$ and 1.11$\times$ in ASIC and FPGA platforms, respectively, compared to a single 8-bit multiplier. Furthermore, the ROM rounding adds negligible area overhead to the 8-bit multiplier while it can significantly improve the classification accuracy, especially fixed-point precision using few number of bits, e.g., 4-bit representation. Moreover, SR consumes more area and power compared to other rounding methods.

Table~\ref{tab:hardware_res} shows the hardware implementation result of DianNao~\cite{chen2014diannao} on an ASIC platform. DiaNao has 16 neuron processing units that processes 16 synapses. So, we implement 256 16-bit multipliers (NFU-1) with 16 adder trees consist of 15 adders (NFU-2) for Conv layers. For pooling layers, we implement 16 multipliers and adders (NFU-3). In these experiments, we employ different rounding methods on RMAC and RMULT approaches. In RMAC, we employ 32-bit adders after performing 16-bit multiplication, and then we adjust the final result with different approaches (at the end of NFU-2). In RMULT, we adjust the precision after performing each multiplication, and then we employ the 16-bit adders to calculate the MAC operation. The results indicate that RMULT employs truncation reduces the area by 1.09$\times$ and power by 1.15$\times$ compared to the RMAC approach. Moreover, the rounding methods add more area and power in RMULT compared to RMAC due to the frequent use of these rounding methods. For instance, RN adds 3\% more area overhead to the truncation method in RMULT, while in RMAC, RN adds a negligible area overhead.

\begin{table}[]
 \caption{ASIC implementation result of DianNao with different configuration}
\label{tab:hardware_res}
\scriptsize
\begin{tabular}{c|c|c|c||c|c|c}
\cmidrule[1pt](lr){1-7}
\multirow{2}{*}[-1.5ex]{\textbf{\begin{tabular}[c]{@{}c@{}} Rounding\\  Method\end{tabular}}} & \multicolumn{3}{c||}{\textbf{RMULT}} & \multicolumn{3}{c}{\textbf{RMAC}} \\ \cmidrule(lr){2-4} \cmidrule(lr){5-7}
 & \textbf{\begin{tabular}[c]{@{}c@{}}Area\\  \textbf{$\mu m^2$ (k)}\end{tabular}} & \textbf{\begin{tabular}[c]{@{}c@{}}Power\\  (mW)\end{tabular}} & \textbf{\begin{tabular}[c]{@{}c@{}}Freq.\\  (MHz)\end{tabular}} & \textbf{\begin{tabular}[c]{@{}c@{}}Area\\  \textbf{$\mu m^2$ (k)}\end{tabular}} & \textbf{\begin{tabular}[c]{@{}c@{}}Power\\  (mW)\end{tabular}} & \textbf{\begin{tabular}[c]{@{}c@{}}Freq.\\  (MHz)\end{tabular}} \\ \cmidrule[1pt](lr){1-7}
\textbf{ Truncation} & 482.88 &	106.41 & 250 & 530.83 & 123.15 & 250   \\ \cmidrule(lr){1-7}
\textbf{ SR} & 533.86 & 119.78 & 250 & 533.891 & 123.45 & 250   \\ \cmidrule(lr){1-7}
\textbf{ RN} & 498.70 & 111.17 & 250 & 532.71 & 123.00 & 250   \\ \cmidrule(lr){1-7}
 \textbf{ ROM} & 483.333 & 106.49 & 250 & 530.81 & 122.36 & 250  \\\cmidrule[1pt](lr){1-7}
\end{tabular}
\end{table}

\begin{table*}[t]
	\centering
	\caption{Compare Previous Study on the CNN Hardware Accelerators Using Fixed-point Representation and this work}
	\label{tab:compare}
	\footnotesize
	\begin{threeparttable}
		\begin{tabular}{c|c|c|c|c|c|c|c|c}
			\cmidrule[1pt]{1-9}
			\textbf{\begin{tabular}[c]{@{}c@{}}Fixed-point \\ Representation\end{tabular}} &
			\textbf{\begin{tabular}[c]{@{}c@{}}Different \\ Precision\end{tabular}} & 
			\textbf{Adjust Methods\tnote{$\diamond$}} & 
			\textbf{\begin{tabular}[c]{@{}c@{}}Rounding \\ Methods\end{tabular}} & 
			\textbf{\begin{tabular}[c]{@{}c@{}}Adjusting \\ Position\end{tabular}} & 
			\textbf{\begin{tabular}[c]{@{}c@{}}Hardware \\ Implementation\end{tabular}} & 
			\textbf{\begin{tabular}[c]{@{}c@{}}Open-source \\ Framework\end{tabular}} &
			\textbf{Re-train\tnote{$\dagger$}} &
			\textbf{\begin{tabular}[c]{@{}c@{}}Dynamic \\ Fixed-point\end{tabular}} \\  \cmidrule[1pt]{1-9}
			\textbf{\cite{hashemi2017understanding}} & \ding{51} & \ding{55} & \ding{55} & \ding{55} & \ding{51} & \ding{55}  & \ding{51}  & \ding{55}\\ 
			\rowcolor{light-gray}\textbf{\cite{lo2018fixed}} & \ding{51} & \ding{55} & \ding{55} & \ding{51} & \ding{51} & \ding{55}  & \ding{51}  & \ding{51}\\ 
			\textbf{Laius~\cite{li2017laius}} & \ding{55} & \ding{55} & \ding{55} & \ding{55} &  \ding{51} & \ding{55}  & \ding{51}  & \ding{55}\\
			\rowcolor{light-gray}\textbf{Ristretto~\cite{gysel2018ristretto}} & \ding{51} & \ding{55} & \ding{55} & \ding{55} & \ding{55} & \ding{51}  & \ding{51}  & \ding{51}\\
			\textbf{\cite{gupta2015deep}\tnote{$\triangle$}} & \ding{51} & \ding{55} & \ding{51} & \ding{55} & \ding{51} & \ding{55}  & \ding{51}  & \ding{55}\\
			\rowcolor{light-gray}\textbf{This work} & \ding{51} &    \ding{51} & \ding{51} & \ding{51} & \ding{51} & \ding{51}\tnote{$\ddagger$}  & \ding{55} & \ding{55} \\ 
			\cmidrule[1pt]{1-9}
		\end{tabular}
		\begin{tablenotes}
			\item[$\diamond$] The selection methods that are introduced in Section~\ref{subsec:operation}.
			\item[$\dagger$] Apply different methods in the training phase to increase the classification accuracy with the fixed-point representation. 
			\item [$\triangle$] This work investigates the impact of different rounding methods on the training phase.
			\item [$\ddagger$] \url{https://github.com/3S-Lab/FixFlow}
		\end{tablenotes}	
	\end{threeparttable}	
\end{table*}

\section{Discussion}
\label{sec:discussion}

This paper introduces a Fixflow framework to evaluate the effect of fixed-point computation on CNN inference comprehensively. To this end, we evaluate the effects of (1) different rounding methods; (2) different approaches to adjust the precision after fixed-point multiplication; (3) different positions that can adjust the precision (RMULT and RMAC) in hardware accelerators on classification accuracy. We provide a hardware implementation for these approaches to have a comprehensive evaluation. 

Our evaluation indicates that rounding methods significantly affect classification accuracy, especially when employing low-precision fixed-point numbers. RN obtains the best accuracy in rounding methods at the cost of area overhead. Furthermore, the ROM rounding method increases the accuracy with negligible area overhead; therefore, it is suitable for resource-constrained applications. Moreover, results show that, although SR is the best rounding method for training~\cite{gupta2015deep}, in the inference phase, SR is not an efficient solution in terms of accuracy and area overhead. 

According to our experimental result (Section~\ref{sec:eval}), in a low-precision fixed-point representation, RMAC is the best approach that can be used in hardware accelerators. Additionally, in a high-precision fixed-point number, the RMULT is the best approach due to low area overhead and low accuracy loss. The NORM method can achieve the best classification accuracy; however, some considerations should be handled in this method, such as selecting MSBs of the result and finding the scale factor for weights and inputs. The classification accuracy drops by 0.6\% when NORM is used in a 4-bit fixed-point representation compared to the floating-point representation for the MNIST dataset without employing any fine-tuning techniques. However, we suggest using the BC method in high-precision fixed-point numbers due to 
its less complex implementation. Table~\ref{tab:compare} summarizes the comparison between this work and related work. Note that the re-train and dynamic fixed-point methods that are used to improve the accuracy are outside the scope of this paper. Here, we want to evaluate the effect of hardware implementation on inference accuracy; therefore, the approaches investigated in this paper are suitable for hardware designer. These methods can be implemented on hardware accelerator to improve the accuracy without any pre-processing stages. 
Dynamic fixed-point adds hardware complexity, and pre-processing stages, while re-training need multiple pre-processing stages and access to the entire dataset to improve the accuracy. Post-training quantization methods do not require an entire dataset or time-consuming re-training; however, these method do not investigate hardware design configuration on inference accuracy.  
Clearly, these methods can significantly improve the accuracy, we believe employing our finding with these methods can further improve inference accuracy. Moreover, we built an open-source framework with various fixed-point configurations mentioned in the paper for the CNN inference phase. We hope that this framework can help other researchers to evaluate fixed-point representation and different arithmetic units, including truncated and approximate multipliers on the CNN hardware accelerators, more efficiently.


\section{Conclusion}
\label{sec:conclusion}
Nowadays, hardware accelerators are commonly used to accelerate CNN inference and training phases. Hardware accelerators improve the efficiency of CNNs in various applications such as IoT. This paper introduces a framework called Fixflow to investigate different approaches to perform fixed-point computation with different precisions. Furthermore, for a comprehensive evaluation, we provide hardware implementation for these approaches. Our findings can be employed in the CNN inference hardware accelerators using fixed-point representation to archive a better accuracy. As future work, we can investigate different approaches for performing the fixed-point computation in the CNN training phase, evaluate dynamic fixed-point and evaluate these configurations on larger datasets.

\bibliographystyle{ieeetr}
\bibliography{ms}

\end{document}